\documentclass{llncs}
\usepackage{mathptmx}       
\usepackage{helvet}         
\usepackage{courier}        
\usepackage{type1cm}        
\usepackage{makeidx}         
\usepackage{graphicx}        
\usepackage{multicol}        
\usepackage[bottom]{footmisc}
\usepackage{subfigure}
\usepackage{amsfonts}
\usepackage{algorithm}
\usepackage[noend]{algpseudocode}

\makeatletter
\def\BState{\State\hskip-\ALG@thistlm}
\makeatother

\usepackage[cmex10]{amsmath}

\begin{document}

\title{Dissecting embedding method: 
learning higher-order structures from data}
%
\author{Liubov Tupikina\inst{1} \inst{2} \inst{3} \and Hritika Kathuria\inst{2}
}
\authorrunning{Author1Name Author1Surname et al.} 

\institute{
Nokia Bell labs, France, Paris, France, 
\and 
LPI, University Paris Descrartes, Learning Planet Institute, INSERM,
Paris, France
\and
MIPT, Moscow Insitute of Physics and Technology, Moscow, Russia \\
}
\maketitle    

\section{Introduction and problem statement}
%
Active area of research in AI is the theory of manifold learning and finding lower-dimensional manifold representation on how we can learn geometry from data for providing better quality curated datasets  \cite{Melas-Kyriazi}. 
There are however various issues with these methods related to finding low-dimensional representation of the data, the so-called curse of dimensionality 
\cite{DeDampierre}. 
Geometric deep learning methods for data learning often include set of assumptions on the geometry of the feature space. 
Some of these assumptions include pre-selected metrics on the feature space, usage of the underlying graph structure, which encodes the data points proximity.
However, the later assumption of using a graph as 
the underlying discrete structure, encodes only the 
binary pairwise relations between data points, restricting ourselves from capturing more complex higher-order relationships, which are often often present in various systems \cite{Melas-Kyriazi}, \cite{Battiston}.  
These assumptions together with data being discrete and finite can cause some generalisations, which are likely to create wrong interpretations of the data and models outputs. Hence overall this can cause wrong outputs of the embedding models themselves, while these models being quite and trained on large corpora of data, such as BERT, Yi and other similar models. 
\\
The objective of our research is twofold, first, it is to develop the alternative framework to characterize the embedding methods dissecting their possible inconsistencies using combinatorial approach of higher-order structures which encode the embedded data. Second objective is to explore the assumption of the underlying structure of embeddings to be graphs, substituting it with the hypergraph and using the hypergraph theory to analyze this structure. We also demonstrate the embedding characterization on the usecase of the arXiv data.  

\begin{figure}
\centering
\includegraphics[width=0.6\textwidth]{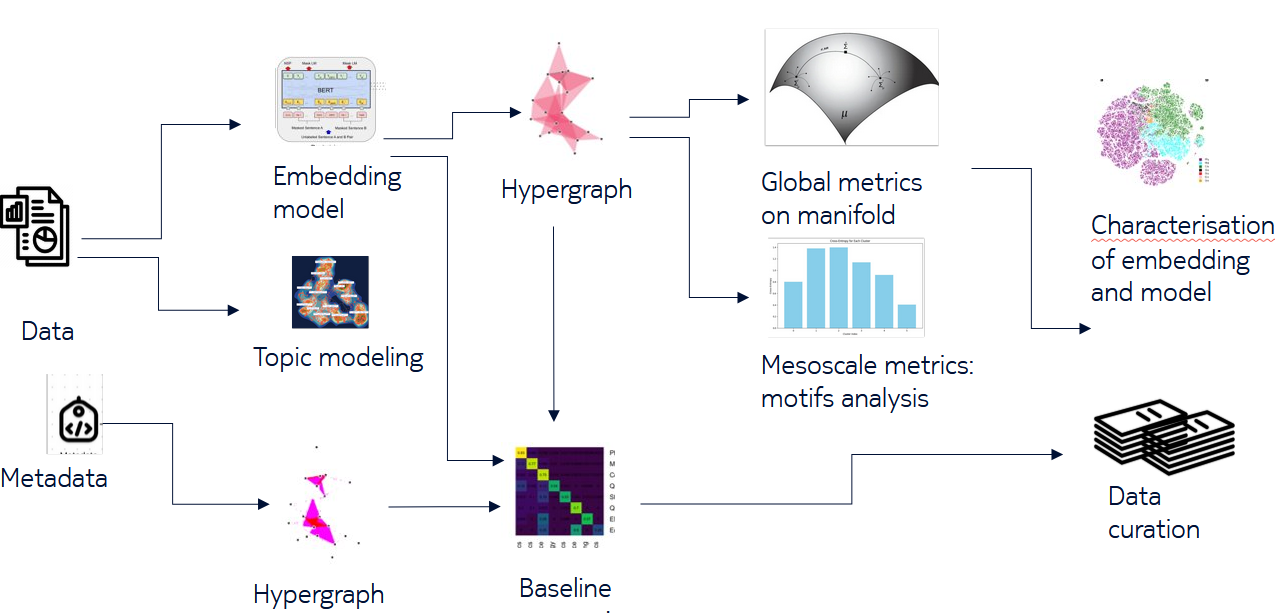}
\caption{ Scheme of the methodological framework. 
}
\label{scheme1}       
\vspace{-0.4cm}
\end{figure}

\section{Data}

The arXiv dataset encompasses scientific articles published online between 1992 and 2018, which span $175$ scientific fields \cite{Singh}. 
Each article has the ground-truth information (metadata) stored as a category or a tag, e.g. \emph{cond-mat.dis-nn} and others. 

\section{Methodological framework}

Our methodological framework consists of the following steps, \textbf{Fig.~\ref{scheme1}}. \textbf{First}, we apply the embedding model 
to the datapoints $D_i$ with the metadata $m_i$ to get the coordinates of the embedded datapoints.
\textbf{Second}, we cluster the datapoints $D_i$ in the embedding using some clustering method (e.g. $k$-means clustering method) 
and apply ML metrics (e.g. confusion matrix, accuracy scores) to deduce scores of the embedding model, which produced the vector representation of datapoints $D_i$ using the metadata labels of clusters $m_i$.
\textbf{Third,} we test the discripancies in the resulted confusion matrix by constructing the \emph{neighborhood hypergraph} $H$ from datapoints $\{D_i\}_{i=1}^n$ and calculate hypergraph motifs in this hypergraph $H$, corresponding to the embedding. Then we analyze the hypergraphs constructed from the metadata $m_i$ and sub-hypergraphs, which contain the misclassified tags based on the metrics from the second step of the method. 
Motifs in hypergraphs are subhypergraphs structures. By studying them one can dissect the most popular of the higher-order intricate properties of data structures  \cite{Zapata}. 
\subsubsection{Neighborhood hypergraph construction method.}
 The \emph{neighborhood hypergraph} $H_n$
 is constructed from data points $\{D_i\}_{i=1}^n$ based on the idea of estimating the relations between the datapoints. 
 If datapoints $(D_{i_1}, ..., D_{i_k})$ are close to each other, they form a hyperedge of a hypergraph in $H_n$ of $k$-arity. 
Depending on the nature of data there are different ways to characterize closseness of datapoints, which can be based on their geometrical or semantical proximity. 
 Geometrical proximity of points $\{D_i\}$ is estimated using typical $l2$ metrics or cosine similarity applied to coordinates of each processed datapoint $D_i$.
 Semantical proximity of points is encoded by labelled similarity of datapoints, for instance, in case of arXiv data it means that
 if paper $\{D_1\}$  has tags $a,b,c$, paper $D_2$ has tags $a,b$ and paper $D_3$ has tags $a,d$, then these papers are encoded as nodes of the hypergraph forming the hyperedge with the field $a$. 
  One can also construct other encoding of the higher-order structure from data using \emph{metadata} $\{m_i\}_{i=1}^m$ information $H_m$ which encodes relations between tags themselves and can characterize the fields closseness instead of using the standard binary graph representation of data as co-tags networks \cite{Singh}: when tags $a,b,c$ are nodes of hypergraph, these nodes share the hyperedge, iff there is at least one paper with all these tags, like paper $D_1$ in this case. 
Our intuition behind using the \emph{neighborhood hypergraph} representation of data is based on the observation that in many textual data similarities between documents are not just based on one measure of closeness but rather on high-order relations like in words: word \emph{France} is close to \emph{Italy} and \emph{Slovenia} in different way than \emph{France} to the word \emph{Paris}. 
\subsubsection{The embedding models.}
In this short paper we choose 
BERT as \textbf{the embedding model}, general autoencoder architecture, which provides the embedding of the textual data \cite{DeDampierre}. Our framework can be applied to other embedding models.

\begin{figure}
\centering
\includegraphics[width=0.4\textwidth]{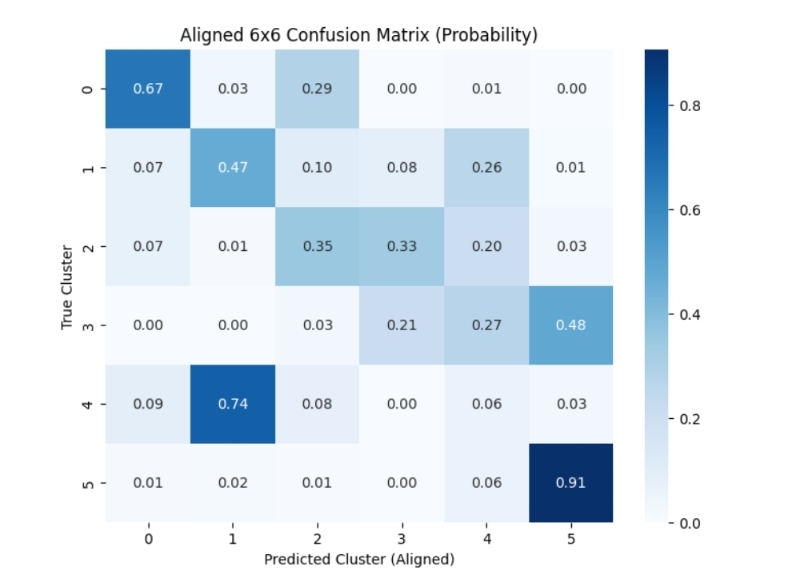}
\includegraphics[width=0.4\textwidth]{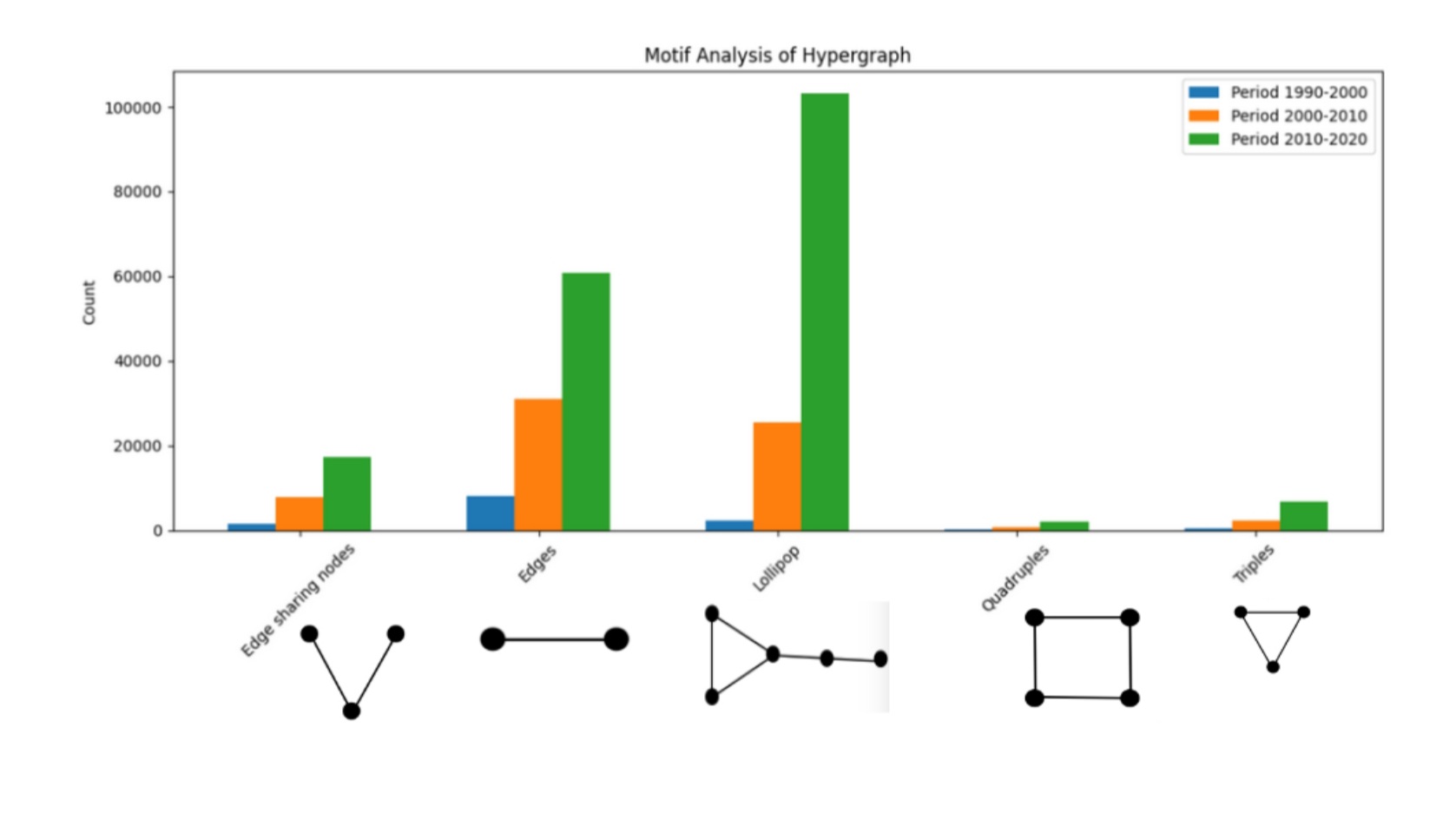}
\caption{(Left) Confusion matrix is calculated for the clustered datapoints from BERT embedding method to compare them to the clusters from ground truth labels (metadata). 
We consider subsample of data from (2000-2011). (Right) Characterisation of persistent hypergraph motifs in the hypergraphs constructed from abstracts datapoints $D_i$ for each decade. 
}
\label{motif}       
\vspace{-0.4cm}
\end{figure}


\subsubsection{Analytical background of the method and general relation to the manifold learning.}
Our general motivation comes from the manifold regularisation method, which allows one to learn the properties of data limiting the space of solutions to the manifold. 
The central idea of the manifold learning is
to form a neighborhood graph $G_n$ \cite{Melas-Kyriazi}.
 We aim further to extend the idea of considering the neighborhood hypergraph $H_n$ instead of the neighborhood graph as the underlying learned structure from data. 
Usually we assume that the feature space of data has the standard $l2$ metric, which induces the Hilbert space.
The neighborhood graphs (hypergraphs) are constructed such that one chooses $k$ nearest points as the neighborhood or points at the maximum distance $r$ for the neighborhood. 
Manifold regularization further simplifies assumptions using a binarized metric or heat kernel.
The problem of the manifold regularisation with the hypergraph $H_n$ then can be further formalized as:
$$ f^* = \min_{f\in h_k}(L(f)) + \lambda (f^T L (H_n) f),$$
where 
$L(H_n)$ is the generalized hypergraph Laplacian.  $(f, L(H_n) f) = f^T L(H_n)f$ is a scalar product, $\lambda$ is the regularization parameter \cite{Melas-Kyriazi}, $f^T$ is the transpose of a vector, where $\min_{f\in h_k}$ is minimum found from the set of possible hypothesis e.g. hypothesis on spaces search. 
Finding the underlying hypergraph can then extend the general method for manifold learning from data, yet it has some drawbacks, such as increasing complexity of the computations.

\section{Results}

We cluster datapoints $D_i$ with coordinates estimated by BERT method. Then we estimate the confusion matrix,  \textbf{Fig. \ref{motif} (left)}, to characterize the potential discrepancy in accuracy of the embedding method, 
 verifying the clustered datapoints according to the labels of metadata $m_i$. 
Most of the misclassified  datapoints belong to the cluster with \emph{condmat} metadata field label, which is also known to be interdisciplinary field \cite{Singh}. Majority of papers in \emph{condmat} typically share common tags with other fields, which correspond to its propensity to be involved in various hyperedges of the \emph{neighrborhood hypergraph} reconstructed from the data. For exploring this aspect we analyze the hypergraph, which encodes the proximity of abstracts (our datapoints $D_i$). For this we study the higher-order motifs in the whole hypergraph and its subhypergraphs, \textbf{Fig. \ref{motif} (right)}. 
We studied both, hypergraph $H_n$ constructed from all datapoints $D_i$, as well as subhypergraphs, which consists of only nodes with \emph{condmat} as their tags. In those subhypergraphs we found higher-order motifs, such as \emph{triples, lollipops} in both structures, and we were able quantitatively characterize them. 
Moreover, we see the persistent growth in number of such higher-order motifs from the first decade of arXiv data to the present one, and specifically lollipop motifs, which are the most persistent patterns in these hypergraphs. 


\section{Conclusions}
In our work we develop the framework to dissect the geometrical properties of embeddings based on the higher-order structures. This approach can be used in addition to traditional combinatorics and information theory based approaches to characterize embeddings \cite{Grootendorst}. 
In particular, in our approach we characterize possible inconsistencies in embedded data estimating the prevailing motifs in subhypergraphs, associated with the misclassified data.
We plan to incorporate hypergraph embedding methods \cite{Evans} further to investigate their applicability to data curation for LLMs training, as well as to dimensionality reduction methods.
Currently we studied only BERT embedding model, yet  
our methodology has the potential to be further extended to differentiate between distinct embeddings, when we construct corresponding hypergraphs to each separate embedding method and then compare embeddings using the hypergraphs theory \cite{Zapata}.


\end{document}